# FineWeb-zhtw: Scalable Curation of Traditional Chinese Text Data from the Web

Cheng-Wei Lin 林承緯 [1*], Wan-Hsuan Hsieh 謝宛軒 [1*], Kai-Xin Guan 關凱欣 [1*],
Chan-Jan Hsu 許湛然 [1], Chia-Chen Kuo 郭嘉眞 [2], Chuan-Lin Lai 賴傳霖 [2],
Chung-Wei Chung 鐘崇偉 [2], Ming-Jen Wang 王明仁 [2], Da-Shan Shiu 許大山 [1]

[1]MediaTek Research    [2]National Applied Research Labatories
{illiamw81, teresa920626, fionag1218}@gmail.com
{cckuo, c00lcl00, 2403047, c00mrw00}@narlabs.org.tw

## Abstract

The quality and size of a pretraining dataset significantly influence the performance of large language models (LLMs). While there have been numerous efforts in the curation of such a dataset for English users, there is a relative lack of similar initiatives for Traditional Chinese. Building upon this foundation of FineWeb (Penedo et al., 2024a), we introduce FineWeb-zhtw, a dataset tailored specifically for Traditional Chinese users. We came up with multiple stages of meticulously designed filters to cater to the linguistic difference between English and Traditional Chinese, to ensure comprehensiveness and quality. We determined effectiveness from querying dataset samples with three main objectives. Our code and datasets are publicly available.

## 1 Introduction

The performance of large language models (LLMs) is significantly influenced by the quality and size of the pretraining datasets. While the exact data recipe remains undisclosed for many chatbot pursuits, it is generally perceived that trillions of high-quality tokens are the cornerstone for creating a generalist language model (Touvron et al., 2023a,b; Meta, 2024; Jiang et al., 2024). The vast expanse of the World Wide Web offers an enormous and diverse reservoir of data, making it an invaluable resource for experiments that require large-scale data. To extract relevant text information from raw HTML content, various preprocessing techniques have been established (Penedo et al., 2023b; Raffel et al., 2023; Gao et al., 2020). Among these methods, FineWeb (Penedo et al., 2024a) has set a new paradigm in this domain, boasting a very competitive data-performance trade-off. Despite the many efforts in English, there remains a notable absence of publicly available, high-quality datasets specifically tailored for Traditional Chinese. Further, the unique linguistic and cultural nuances of Traditional Chinese, such as lack of spacing, hinder the direct projection of current pipelines, which necessitates a specialized refinement approach.

In response to this need, we introduce FineWeb-zhtw, a dataset meticulously curated to provide a robust and comprehensive collection of Traditional Chinese text catered to the language usage of the Taiwanese community. By employing advanced filtering techniques, we ensure high quality and relevance on the samples of FineWeb-zhtw.

The creation of FineWeb-zhtw involves several stages of filtering, including basic filtering, multi-staged language identification, followed by Gopher (Rae et al., 2022), C4 (Raffel et al., 2023), and FineWeb (Penedo et al., 2024a) quality filters. Each of these stages is designed to enhance the dataset's overall quality and relevance. For the implementation of these filters and the curation process, we utilized `datatrove` (Penedo et al., 2024b), a tool developed by the HuggingFace team that facilitated our efforts effectively. We then evaluate the dataset samples on three custom metrics to understand the effectiveness of our filtering pipeline.

Through FineWeb-zhtw, we aim to advance the development of natural language research for Traditional Chinese, providing a crucial re-

---

[*] Equal Contribution. Work done during internship at MediaTek Research. 林承緯 is studying at the Department of Computer Science of National Yang Ming Chiao Tung University, 謝宛軒 and 關凱欣 are studying at the Department of Information Management of National Taiwan University

source for the community to build upon.

## 2 Building FineWeb-zhtw

The raw data for FineWeb-zhtw is sourced from Common Crawl, an extensive web archive that captures a comprehensive snapshot of the internet. The data is stored in Web ARChive (WARC) format, which is a standard format for web-crawled data. This format includes both the retrieved content and metadata, making it suitable for large-scale data processing tasks. For the parameters mentioned in the paragraph, we determined the parameters using grid search by manually inspecting the quality of filtered-in and filtered-out data.

### 2.1 Basic filtering and Text Extraction

The initial stage of building the FineWeb-TC revolves around extracting text from html webpages. We use the `trafilatura` tool favoring precision to remove potential headers and footers. Since html cleaning is an expensive operation compared to filtering out entire webpages, we implemented a pre-filter that excludes documents that are guaranteed not to be in Traditional Chinese to minimize computational cost. The pre-filter defines fuzzy Traditional Chinese tokens by an unicode range[1], and any document without 5 consecutive characters in the set will be filtered out.

We also applied a URL filter, which blocks flagged websites that have low content value such as spam.

### 2.2 Language Identification

The filtering process for FineWeb-TC is to identify core text from html that are naturally written in Traditional Chinese. Most language identifiers, such as fasttext (Bojanowski et al., 2017), do not differentiate between Traditional Chinese and Simplified Chinese. As we could not find a pre-defined document-wise filter with satisfying accuracy on these two variants of Chinese, we designed a custom filter in addition to fasttext, with character and word-phrase filtering. We remove documents that match any of the specified phrases, prioritizing the precision of the remaining entries. This step is crucial for ensuring that the dataset accurately reflects and aligns with the language and preferences of the local community.

### 2.3 Gopher filters

Next, we applied the Gopher Quality Filter, which uses heuristic rules to ensure the text content adheres to specific quality standards. This custom filter evaluates various textual characteristics and excludes documents that do not meet the established criteria.

- **Document Length**: Short documents often lack useful information, while excessively long documents are more likely to be spam and can skew the dataset. Therefore, we excluded documents with fewer than 50 words or more than 100,000 words.

- **Symbol-to-Word Ratio**: Excessive use of symbols can indicate low-quality content. To filter these out, we excluded documents with a symbol-to-word ratio exceeding 0.1.

- **Ellipsis Lines Ratio**: High frequencies of ellipses can suggest truncated content. These articles with unfinished sentences decrease in value tremendously under the auto-regressive modeling objective. Documents with more than 30% of lines ending with ellipses were excluded.

- **Stop Words Count**: Documents lacking sufficient stop words often do not contain expressive and coherent sentences. Therefore, we excluded those without any of the predefined stop words.

By excluding documents that fail to meet these criteria, the Gopher Quality Filter ensures a dataset with more consistent and high-quality content suitable for analysis and application.

### 2.4 C4 filters

To further refine the FineWeb-zhtw dataset, we implemented the C4 Quality Filter and customized its parameters to optimize for filtering Traditional Chinese content. Besides document-wise filtering, the C4 filter also highlights quality checks on each line of the documents. Uninformative lines are removed to

---
[1] $\backslash u3040 - \backslash u3090 \backslash u30a0 - \backslash u30ff \backslash u4e00 - \backslash u9fff$

increase the density of relative information in the data.

The C4 Quality Filter uses several key criteria to determine the quality of a line in the content:

- **Filter JavaScript**: Lines containing the word `javascript` or curly brackets are often code snippets or irrelevant content. These lines are excluded to keep the dataset focused on textual content.

- **Filter Policy-Related Substrings**: Lines containing policy-related substrings such as `terms of use`, `privacy policy`, or `cookie policy` are usually boilerplate text. Excluding these lines helps remove non-informative content.

- **Bracket Ratio**: A high bracket ratio can indicate the presence of excessive code or other structured content. Documents with a bracket ratio exceeding 0.01 are excluded. The bracket ratio is calculated as the proportion of brackets in the document.

By applying these criteria, the FineWeb-zhtw dataset is cleaner, more relevant, and better suited for subsequent analysis and applications.

## 2.5 FineWeb quality filters

Finally, we implemented the FineWeb Quality Filter and tuned its parameters to enhance the quality of Traditional Chinese content. This custom filter ensures that the text content meets high-quality standards by analyzing key textual characteristics and excluding documents that do not meet these criteria.

The FineWeb Quality Filter uses the following criteria to assess document quality:

- **Line Punctuation Ratio**: Proper punctuation is essential for text coherence. We exclude documents with a line punctuation ratio below 0.04 to ensure that only well-punctuated, readable content is included.

- **Short Line Ratio**: Documents with a high ratio of lines shorter than 10 characters are likely fragmented. By excluding documents where this ratio exceeds 0.8, we ensure that the dataset contains substantial and informative content.

- **Character Duplication Ratio**: Excessive repetition of characters can indicate low-quality content. We exclude documents with a duplication ratio above 0.3 to filter out repetitive text and retain diverse content.

- **New Line Ratio**: A high new line ratio suggests poor text structure. We exclude documents with a new line ratio greater than 0.3 to maintain well-organized and coherent documents.

By applying these criteria, the amount of low-quality or poorly structured content in the dataset is significantly reduced.

## 2.6 Minhash Deduplication

We first applied a document-level minhash deduplication to remove repetitive content. Then, we iteratively removed lines of content leading or trailing the document, if that line has over 100 occurrences in one dump of Common Crawl.

## 2.7 The final FineWeb-zhtw dataset

We have implemented a comprehensive pipeline to process Common Crawl data, utilizing various filtering techniques at each stage. The process begins with extracting documents from WARC files. Then, filters mentioned in the sections above are applied sequentially, starting with the basic filtering (Section 2.1), followed by the Language Identification (Section 2.2). Next, we use the gopher filters (Section 2.3), C4 filters (Section 2.4), FineWeb quality filters (Section 2.5), and finally, minhash deduplication (Section 2.6).

Figure 1 illustrates the global removal rate and the relative retention rates of documents at each stage, offering not only an overview of the total Traditional Chinese data reserve but also a comparative analysis of the impact of each filter. The final screening rate is around 99.5%, leaving 0.5% of effective Traditional Chinese data available.[2]

---

[2] 14.04 gigabytes of pure text data remain for the dump CC-MAIN-2024-26

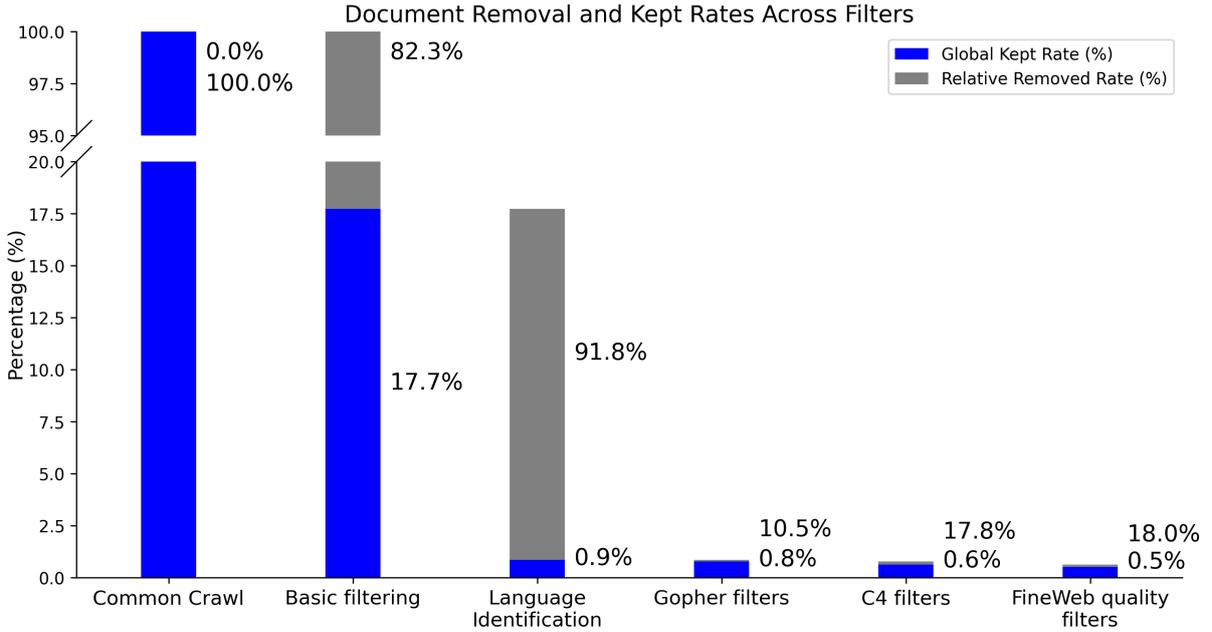

Figure 1: **Relative removal rate and global kept rate of Common Crawl during each filtering stage.** We report the removal rate with respect to each previous stage, and the overall kept rate (Penedo et al., 2023a). Rates are calculated by the number of documents until the language identification stage. For later line-by-line filtering, rates are measure in tokens (bytes).

## 3 Evaluation

### 3.1 Evaluation Methodology

To ensure the quality and relevance of the FineWeb-zhtw dataset, we evaluated the final dataset using the following criteria:

1. **Traditional Chinese and Language Naturalness**: Accurate use of Traditional Chinese characters and natural language is essential for accessibility and usability by native speakers. Proper language usage enhances readability and comprehension, which are critical for any linguistic dataset.

    This criterion assessed whether the content is written in Traditional Chinese, employs accurate characters, and adheres to Traditional Chinese grammar and conventions. Factors include grammatical correctness, character usage, contextual appropriateness, clarity, and logical structure.

2. **Educational Value**: High educational value ensures that the content contains relevant knowledge. Datasets with great educational content provide reliable materials that enhance knowledge-based question answering of the language model.

    This criterion evaluated whether the content is informative and relevant for learning purposes, considering educational coverage, coherence, structure, and overall value.

3. **Sensitive Content**: Excluding sensitive or inappropriate content is vital for ethical development and ensuring user safety.

    This criterion checked for sensitive or inappropriate content, including violence, pornography, discrimination, and privacy violations.

Each document was rated on a scale from 0 to 5 for each criterion, and the scores were summed to provide a total score for each document.

We utilized "language model as a scoring agent" (Chiang and Lee, 2023; Chiang et al., 2024) to automate and streamline this evaluation. The assistant was given a detailed prompt outlining the criteria and scoring methodology (See Appendix for details), ensuring consistent and accurate assessment.

We randomly selected 1,000 documents from different stages of the pipeline and ran evaluations on the metrics mentioned above, using GPT3.5 on the DaVinci[3] platform.

### 3.2 Evaluation Results

To assess the effectiveness of our filtering process and the quality of the FineWeb-zhtw dataset, we compared it against two ablations: the dataset after basic filtering and the dataset after language identification.

Figure 2 presents the evaluation results for the different datasets, highlighting the distribution of scores and mean values for each criterion. We conducted a t-test on the null hypothesis ($H_0$) that the scores are identical. Based on the resulting t-statistic and p-values, we rejected the null hypothesis for p-values less than 0.05. Results show that the FineWeb-zhtw dataset has consistent improvements across all categories compared to the datasets after basic filtering and language identification.

In the **Traditional Chinese and Language Naturalness** category, FineWeb-zhtw led with a mean score of 2.42, compared to 2.19 for the dataset after language identification and 1.72 for basic filtering. Significant improvements with FineWeb-zhtw are supported by a t-statistic of 5.36 (p-value = 9.14e-08) and for the dataset after language identification over basic filtering by t-statistic = 10.13 (p-value = 1.53e-23). The significant difference between FineWeb-zhtw and basic filtering is highlighted by a t-statistic of 15.23 (p-value = 1.29e-49).

For the criterion **Educational Value**, FineWeb-zhtw scored a mean of 2.04, outperforming the dataset after language identification (1.77) and basic filtering (1.54). Significant differences are evident with FineWeb-zhtw compared to the dataset after language identification (t-statistic = 4.96, p-value = 7.56e-07) and the dataset after language identification compared to basic filtering (t-statistic = 4.12, p-value = 3.93e-05). The difference between FineWeb-zhtw and basic filtering is supported by a t-statistic of 8.78 (p-value = 3.38e-18).

In the **Sensitive Content** category, as the score is already high, no statistically signifi-cant gains are achieved from the FineWeb filtering stage. These scores reveal that sensitive content is not a huge concern when using CommonCrawl data under our pipeline.

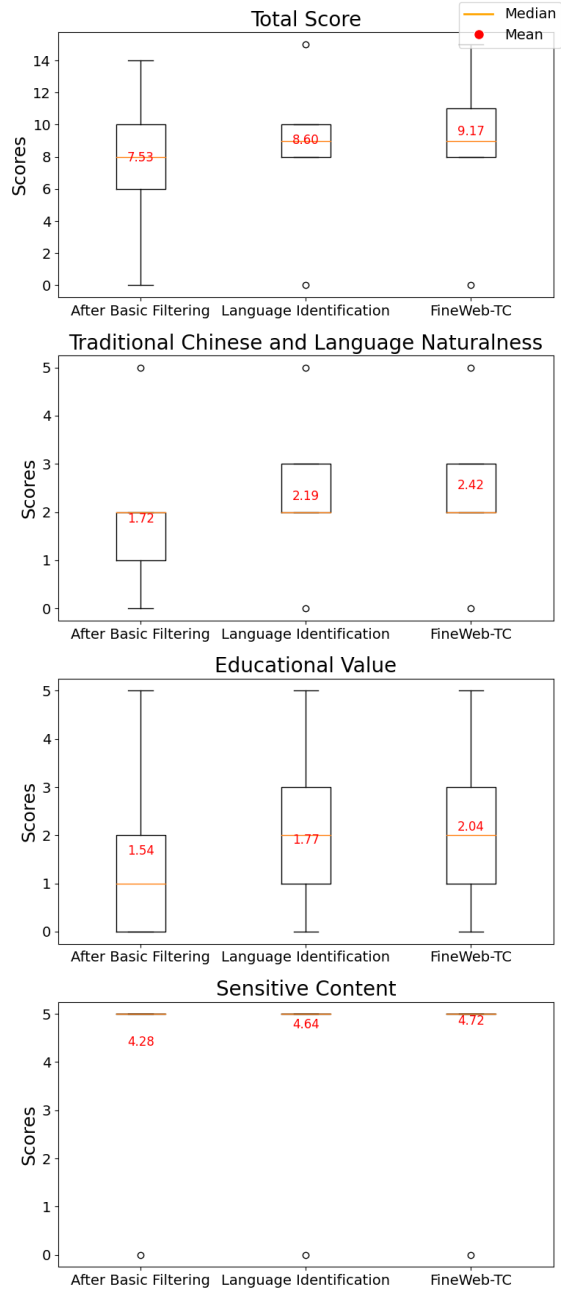

Figure 2: Comparison of evaluation results across different datasets. The FineWeb-zhtw dataset demonstrates significant improvements in Total Score, Traditional Chinese and Language Naturalness, and Educational Value compared to the dataset after language identification and basic filtering, with statistically significant differences (as indicated by t-test results).

---

[3] https://dvcbot.net/

Finally, for the **Total Score**, FineWeb-zhtw achieved a mean of 9.17, higher than 8.60 for the dataset after language identification and 7.53 after basic filtering. T-tests confirm significant differences between all pairings.

These results highlight that our sophisticated filtering pipeline is effective at extracting crucial assets that have a higher educational value and language coherence, which have shown to benefit language model training (Penedo et al., 2024a).

## 4 Discussion

Through the realization of FineWeb-zhtw, we have also observed a significant gap between readily available English and Traditional Chinese data. At the language identification stage, the gap between the data size of English and Traditional Chinese is already around 40x. From our empirical screening rate, it is estimated that even if the entire Common Crawl data is completely extracted to train a Traditional Chinese LLM with 70B parameters, the amount would not be sufficient under the Chinchilla scaling law (Hoffmann et al., 2022). It is therefore of great importance that additional efforts be made to curate publicly available data beyond what is found in Common Crawl.

## 5 Conclusion

In summary, the FineWeb-TC dataset, resulting from the advanced filtering process, represents a significant advancement in the field of readily available Traditional Chinese language modeling corpus. The advanced techniques applied have successfully improved the dataset's overall quality, ensuring better adherence to language standards, educational value, and content appropriateness. These findings affirm the efficacy of our comprehensive filtering pipeline in producing a high-quality Traditional Chinese dataset. Finally, we highlight the qualitative differences between English and Traditional Chinese FineWeb that arise from quantitative changes.

## Acknowledgements

We appreciate the accelerative enhancement made by the members of the project titled 'Smart Decision-Making Mobile Platform for Social Workers' (No. M1304129) in the study design.

# Appendix

## Evaluation Details

We show here the prompt that we use for scoring the data samples:

> 評量標準
>
> 1. 繁體中文與語言自然性：內容是否以繁體中文寫成，並使用正確的繁體中文字符；文本是否自然流暢，符合繁體中文的語法和用詞習慣，並且易於閱讀和理解、內容完整。滿分 5 分，評分時請考慮以下因素：
>    - 若語法正確，句子結構自然，得 1 分。
>    - 若使用正確的繁體中文字符，得 1 分。
>    - 若用詞符合繁體中文的習慣，得 1 分。
>    - 若句子簡潔明瞭，易於理解，得 1 分。
>    - 若內容有邏輯性，有頭有尾，得

1 分。

2. 教育價值：內容是否具有值得學習的正面價值。滿分 5 分，評分時請考慮以下因素：

   - 若內容提供與教育話題相關的基本資訊，即使其中包含一些無關或非學術的內容（如廣告和促銷），得 1 分。
   - 若內容涉及某些教育相關元素，但不完全符合教育標準，得 1 分。它可能會混合教育內容和非教育教材，提供潛在有用主題的概述，或者以無條理和不連貫的寫作風格呈現訊息。
   - 若內容適用於教育用途並介紹與學校課程相關的關鍵概念，得 1 分。它是連貫的，儘管可能不全面或包含一些無關的內容。它可能類似於教科書的介紹部分或適合學習但有顯著局限性的基礎教程，比如將概念處理得對於初中學生來說過於複雜。
   - 若內容對於小學或初中等級的教育目的是高度相關且有益的，得 1 分。它展現出清晰和一致的寫作風格，可能類似於教科書的一章或教程，提供大量的教育內容，包括練習和解答，且幾乎沒有無關的內容，概念不會太先進而超出初中學生的理解範圍。內容連貫、集中且有價值的結構化學習。
   - 若內容在教育價值上是出類拔萃的，完全適合在小學或初中教學，得 1 分。它遵循詳細的推理，寫作風格易於理解，提供深入而透徹的主題見解，且沒有任何非教育或複雜內容。

3. 敏感內容：是否包含敏感或不適當的內容。滿分 5 分，評分時請考慮以下因素：

   - 若不包含暴力言論、行爲、與內容，不宣揚言語、肢體等任一形式暴力，不宣揚槍枝與血腥等內容，得 1 分。
   - 若不包含色情言論、行爲、與內容，得 1 分。
   - 若不包含歧視言論、行爲、與內容，不貶低、侮辱或仇恨任一種族、國家、族群、與個人，得 1 分。
   - 若不涉及政治和宗教等敏感話題，不以言論、行爲等任一方式支持或反對任一政黨或教派，得 1 分。
   - 若不包含侵犯隱私或個人權利的言論、行爲、與內容，得 1 分。

評分格式

請按照以下格式提供評分，分數應以整數型態表示，並將每個標準所得分數加總，計算總分：
1. 繁體中文與語言自然性：< 分數 >
2. 教育價值：< 分數 >
3. 敏感內容：< 分數 >
總分：< 總分 >

評分範例

請對以下文本進行評分：
< 待評估的文本 >